\documentclass[letterpaper, 10 pt, conference]{ieeeconf}  

\IEEEoverridecommandlockouts                              

\IEEEoverridecommandlockouts
\usepackage{mathrsfs,amsmath}
\usepackage{graphics} 
\usepackage{epsfig} 
\usepackage{mathptmx} 
\usepackage{times} 
\usepackage{amssymb}  

\usepackage{caption}
\usepackage{cite}
\usepackage{color}
\usepackage{psfrag}
\usepackage{latexsym, amsmath, subfigure, color, amsfonts, amssymb,graphicx}
\usepackage{algorithm}
\usepackage{algorithmic}
\usepackage{cases}
\usepackage{acronym}
\usepackage{subfigure}
\usepackage{amsfonts}
\usepackage{setspace}
\usepackage{epstopdf}

\usepackage{algorithmic}

\begin{document}
	
	\title{\LARGE \bf Real-time 3D Human Tracking for Mobile Robots with Multisensors}
	\author{Mengmeng Wang$^{1}$, Daobilige Su$^{2}$, Lei Shi$^{2}$, Yong Liu$^{1}$ and Jaime Valls Miro$^{2}$
		\thanks{$^{1}$Mengmeng Wang and Yong Liu are with the State Key Laboratory of Institute of Cyber-Systems and Control, Zhejiang University, Zhejiang, 310027, China. Yong Liu is the corresponding author of this paper,	email: yongliu@iipc.zju.edu.cn}%
		\thanks{$^{2}$Daobilige Su, Lei Shi,  and Jaime Valls Miro are with the Centre for Autonomous Systems (CAS), The University of Technology, Sydney, Australia.}%
	}
\maketitle
\begin{abstract}
Acquiring the accurate 3-D position of a target person around a robot provides fundamental and valuable information that is applicable to a wide range of robotic tasks, including home service, navigation and entertainment. This paper presents a real-time robotic 3-D human tracking system which combines a monocular camera with an ultrasonic sensor by the extended Kalman filter (EKF). The proposed system consists of three sub-modules: monocular camera sensor tracking model, ultrasonic sensor tracking model and multi-sensor fusion. An improved visual tracking algorithm is presented to provide partial location estimation (2-D). The algorithm is designed to overcome severe occlusions, scale variation, target missing and achieve robust re-detection. The scale accuracy is further enhanced by the estimated 3-D information. An ultrasonic sensor array is employed to provide the range information from the target person to the robot and Gaussian Process Regression is used for partial location estimation (2-D). EKF is adopted to sequentially process multiple, heterogeneous measurements arriving in an asynchronous order from the vision sensor and the ultrasonic sensor separately. In the experiments, the proposed tracking system is tested in both simulation platform and actual mobile robot for various indoor and outdoor scenes. The experimental results show the superior performance of the 3-D tracking system in terms of both the accuracy and robustness.
\end{abstract}

\section{Introduction}
Tracking people in 3-D is a key ability for robots to effectively interact with humans. It is an essential building block of many advanced applications in the robotic areas such as human-computer interaction, robot navigation, mobile robot obstacle avoidance, service robots and industrial robots. For example, a service robot tracks a specific person in order to provide certain services or to accomplish other tasks in office buildings, museums, hospital environments, or in shopping centers. It is crucial to estimate the accurate positions of the target continuously for subsequent actions. To track the target people across various complex environments, robots need to localize the target and discriminate him/her from other people. In this context, localizing and tracking a moving  target become critical and challenging for many indoor and outdoor robotic applications \cite{gritti2014kinect, knoop2006sensor, bellotto2009multisensor}.

Target tracking for mobile robots has been a popular research topic in recent years, and plenty of methods using various sensors have been developed  \cite{huang2007heterogeneous,jean2013development,kobilarov2006people,bellotto2009multisensor}. Among them, visual tracking enjoys a good population. It is an extremely active research area in computer vision community and obtains significant progress over the past decade \cite{kalal2012tracking,henriques2015high,danelljan2014accurate,germa2010vision,wang2015robust}. However, a monocular camera sensor is limited in providing the 2-D position because it is insufficient to measure the range information from the robot to the target. To introduce range information while retaining the advantages of visual tracking, an intuitive solution is to incorporate heterogeneous data from other sensors \cite{cui2008multi, kobilarov2006people, jean2013development}.

In this paper, we propose a new method for tracking the 3-D positions of a person by multi-sensors in both indoor and outdoor environments with a robotic platform. Due to the reliability and simplicity of the ultrasonic sensors, we fuse the partial position estimation from a camera and an ultrasonic sensor sequentially and exploit their respective advantages. Visual tracking processes videos captured from the camera sensor to estimate the target's locations in the image coordinate. Ultrasonic array sensor offers the range information of the target in the robot coordinate. The actual 3-D positions are estimated by merging these two heterogeneous information sources. This sensor configuration is an alternative to more complex and costly 3-D human tracking systems for mobile robots. Above all, the contributions of our method are summarized as follows:
\begin{enumerate}
  \item An accurate 3-D human tracking system is proposed by fusing a vision sensor with an ultrasonic array sensor sequentially by the extended Kalman filter (EKF);
  \item An improved online visual tracking algorithm is presented to handle the situations of severe occlusion, object missing and re-detection;
  \item The estimated 3-D information is further exploited to improve the scale accuracy of the target in the image coordinate;
\end{enumerate}

In the experiment, we demonstrate the proposed method with both simulation and actual robot platform. The experimental results show that our method performs accurately and robustly in the 3-D human tracking for several challenging conditions such as occlusions, background clutters, scale variations and even when the target is totally missing.

\section{Related Work}
Our work is related to some specific research areas in computer vision and robotics, which are visual tracking, ultrasonic tracking and 3-D location estimation. We will give a brief exposition for each of them in this section.
\subsection{Visual Tracking using Monocular Camera}
Numerous visual tracking algorithms have been developed over the past few decades\cite{kalal2012tracking}\cite{wang2015robust}\cite{henriques2015high}\cite{ma2015hierarchical}. Recently, a group of correlation filter based discriminative trackers have made remarkable improvement in visual tracking field \cite{henriques2015high}\cite{danelljan2014adaptive}\cite{danelljan2014accurate}\cite{liu2016structural}\cite{liu2015real}. Considering the nature of visual tracking, the correlation filter can be solved in the Discrete Fourier Transform (DFT) effectively and efficiently. These methods are excellent in many environments but they are not suitable for the 3-D human tracking in a robot platform because they are not robust enough in the situations of severe occlusions and object missing. In this paper, an improved correlation filter based visual tracking algorithm was developed to provide enhanced robustness and performance in the application of mobile robots.

An exhaustive analysis is beyond this work. Thus we recommend \cite{wu2015object} and \cite{kristan2015visual} for a full understanding about this problem. 
\subsection{Ultrasonic Tracking}
Ultrasonic sensors have been used extensively as time-of-flight range sensors in localizing the tracking targets \cite{mahapatra2008ultra,ullah2012integrated,su2014ultrasonic}. However, one disadvantage of this type of sensor is that when the target moves at the vertical direction of the sonar beam, the calculated locations are usually inaccurate. Another problem with sonar sensors is the reflection from obstacles in the environment will usually cause invalid and incorrect results. Furthermore, relying on the time-of-flight measurement only, the receiver is unlikely able to discriminate multiple sources which means that the system does not work when multiple targets present.

\subsection{Robotic 3-D Human Tracking}
There are a number of different techniques to track persons with mobile robots. Using laser range finders with cameras for person tracking is an option in the robotics community \cite{cui2008multi, kobilarov2006people, bellotto2009multisensor}. Laser scans could be used to detect the human legs at a fixed height. However, this can not provide robust features for discriminating the different persons in the robot's vicinity, thus the detector always tends to fail when one leg occludes the other.

Combining the sonar sensors with cameras is another popular research direction \cite{wilhelm2002sensor,huang2007heterogeneous,jean2013development}. They usually use the sonar sensors to detect the regions that might contain the target in the sonar's field of view. Corresponding regions in the images are then used as additional measurements.This method may be invalid when the ultrasonic sensors lose the target, leading to the fact that the target is beyond the view of the camera.

3-D features from 3-D point clouds reconstructed by RGB-D sensors are used for 3-D human tracking as well \cite{knoop2006sensor, dondrup2015real, gritti2014kinect, munaro2014fast}. However, the minimum distance requirement, narrow field of view, and sensitivity to the illumination variations of the RGB-D sensors limit this technique for robust human tracking applications.
\section{Method}
The proposed 3-D tracking system can be decomposed into three sub-modules, monocular camera sensor tracking model, ultrasonic sensor tracking model and multi-sensor fusion. In this section, the details about these three sub-modules are presented.

\begin{figure}[!t]
	\centering
	\includegraphics[height=3in,width=2.2in,angle=0]{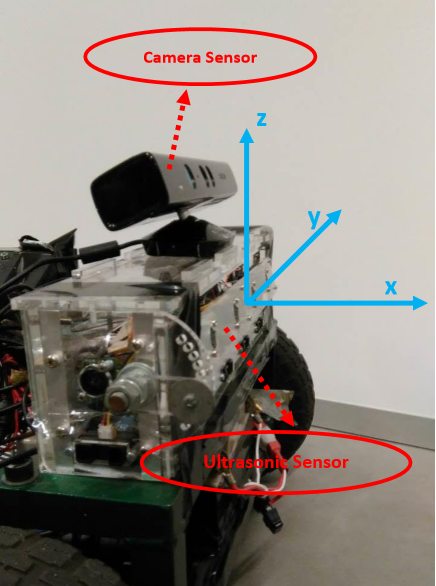}
	\caption{The local robot coordinate. We employ Kinect XBOX 360 to acquire the ground truth of the 3-D position of the target during the tracking process in the experiment. Simultaneously, the RGB camera of Kinect is used as our monocular camera sensor for convenience.}
	\label{coordinate}
\end{figure}

The state of the target ${{\bf{x}}_k} = {\left[ {{x_k},{y_k},{z_k}} \right]^T}$ is defined as the location of the sonar emitter wore by the target person. Here, the subscript $k$ represents the $k$-th time instant. All the variables are defined in the robot's local coordinate frame as shown in Fig.\ref{coordinate}. The target people is sensed by both vision sensor and ultrasonic sensor. To estimate the 3-D position of the people, data acquired from the two sensors are fused sequentially using an EKF.
\subsection{Monocular Camera Sensor Tracking Model}
The monocular camera is installed on the top of the ultrasonic array sensor which is attached on the mobile robots. The vision sensor measurement model ${{\bf{h}}_C}\left( {{{\bf{x}}_k}} \right)$ is a simple camera projection model as shown in Eq. \ref{camera_measure}. 
\begin{subequations}\label{camera_measure}
	\begin{align}
	{{\bf{h}}_C}\left( {{{\bf{x}}_k}} \right) &= {\left[ {{u_{Ck}},{v_{Ck}}} \right]^T} \label{camera_measure:A}\\
	\left[ {\begin{matrix}
		{{u_{Ck}}}  \\ 
		{{v_{Ck}}}  \\ 
		1  \\ 	
		\end{matrix} } \right] &= {\bf{A}}\left[ {\left. {\bf{R}} \right|{\bf{t}}} \right]\left[ {\begin{matrix}
		{{{\bf{x}}_k}}  \\ 
		1  \\ 	
		\end{matrix} } \right]\label{camera_measure:B}
\end{align}
\end{subequations}
where $\left( {{u_{Ck}},{v_{Ck}}} \right)$ is the target's location in the image coordinate, which is estimated by our visual tracking algorithm, $\left[ {\left. {\bf{R}} \right|{\bf{t}}} \right]$ and ${\bf{A}}$ are the extrinsic and intrinsic parameter matrices of the camera respectively. 

For conventional visual tracking, the target is given in the first frame either from human annotation or certain detector. In the proposed 3-D human tracking system, the initial bounding box is calculated by the 3-D to 2-D projection with the target people's height $h$ and the initial 3-D position ${{\bf{x}}_{init}}$. Additionally, we assume the average width of a person is 0.4 meters and the distance from the sonar emitter to the people's feet is $50\%$ of his/her height $h$ in all experiments. Then the initial 3-D positions of left boundary ${{\bf{x}}_{linit}}$, right boundary ${{\bf{x}}_{rinit}}$, head ${{\bf{x}}_{hinit}}$ and feet ${{\bf{x}}_{finit}}$ of the target can be calculated by
\begin{subequations}\label{initial_def}
	\begin{align}
	{{\bf{x}}_{linit}} &= {{\bf{x}}_{init}} + {\left[ {0,0.4/2,0} \right]^T}\label{initial_def:A}\\
	{{\bf{x}}_{rinit}} &= {{\bf{x}}_{init}} + {\left[ {0, - 0.4/2,0} \right]^T}\label{initial_def:B}\\
		{{\bf{x}}_{hinit}} &= {{\bf{x}}_{init}} + {\left[ {0,0,0.5h} \right]^T} \label{initial_def:C} \\
		{{\bf{x}}_{finit}} &= {{\bf{x}}_{init}} + {\left[ {0,0, - 0.5h} \right]^T} \label{initial_def:D}
	\end{align}
\end{subequations}   

The initial width ${w_{init}}$, height ${h_{init}}$ and the center position of the target's bounding box $\left( {{u_{init}},{v_{init}}} \right)$ in the image is calculated as
\begin{subequations}\label{initial}
	\begin{align}
	{w_{init}} &= {u_{rinit}} - {u_{linit}}\label{initial:A}\\
	{h_{init}} &= {v_{finit}} - {v_{hinit}}\label{initial:B}\\
	{u_{init}} &= \left( {{u_{linit}} + {u_{rinit}}} \right)/2\label{initial:C}\\
	{v_{init}} &= \left( {{v_{finit}} + {v_{hinit}}} \right)/2\label{initial:D}
	\end{align}
\end{subequations}
where ${u_{rinit}}, {u_{linit}}$ are the $u$ axis values in the image coordinate of ${{\bf{x}}_{linit}}$ and ${{\bf{x}}_{rinit}}$ calculated by Eq.\ref{camera_measure:B}. Similarly, ${v_{finit}}$ and ${v_{hinit}}$ are the $v$ axis values in the image coordinate of ${{\bf{x}}_{hinit}}$ and ${{\bf{x}}_{finit}}$. 

The presented visual tracking algorithm is based on the Kernelized Correlation Filter (KCF)\cite{henriques2015high} tracker. We extend it with a novel criterion to evaluate the performance of the tracking results and develop a new scale estimation method which estimates the scale variations by combining the projection from the 3-D target position into the 2-D image coordinates with the visual scale estimations.

\subsubsection{KCF Tracker}
In this section, a brief exposition of KCF tracking algorithm is presented, which is described detailedly in \cite{henriques2015high}. The goal is to learn an online correlation filter from plenty of training samples of size $W \times H$. KCF considers all cyclic shifts ${{\bf{s}}_{w,h}}$, $\left( {w,h} \right) \in \left\{ {0,...,W - 1} \right\} \times \left\{ {0,...,H - 1} \right\}$ around the target as training examples. The desired correlation output $y_{w,h}$ is constructed as a Gaussian function with its peak located at the target center and smoothly decayed to 0 for any other shifts. 

The optimal correlation filter $\bf{w}$ is obtained by a function which minimizes the squared error over samples ${{\bf{s}}_{w,h}}$ and their regression labels $y_{w,h}$,
\begin{equation}\label{squared_error}
\mathop {\min }\limits_{\bf{w}} {\sum\limits_{w,h} {\left| {\left\langle {\varphi \left( {{{\bf{s}}_{w,h}}} \right),{\bf{w}}} \right\rangle  - {y_{w,h}}} \right|} ^2} + \lambda {\left\| {\bf{w}} \right\|^2}
\end{equation}
where $\varphi$ denotes the mapping to non-linear feature space with kernel $\kappa$ and the dot-products of $\bf{s}$ and $\bf{s}'$ is $\left\langle {\varphi \left( {\bf{s}} \right),\varphi \left( {{\bf{s}}'} \right)} \right\rangle  = \kappa \left( {{\bf{s}},{\bf{s}}'} \right)$. $\lambda$ is a regularization parameter that controls overfitting. 

With the fact that all circulant matrices are made diagonal by the DFT and some circulant kernels, the solution $\bf{w}$ can be represented as ${\bf{w}} = \sum\limits_{w,h} {{\alpha _{w,h}}\varphi \left( {{{\bf{s}}_{w,h}}} \right)} $, then the optimization goal is the variable $\bf{\alpha}$ rather than $\bf{w}$. 
\begin{equation}\label{alpha}
{\bf{\alpha }} = {{\mathcal F}^{{\text{ - }}1}}\left( {\frac{{{\mathcal F}\left( {\bf{y}} \right)}}{{{\mathcal F}\left( {{{\bf{k}}^{{\bf{ss}}}}} \right) + \lambda }}} \right)
\end{equation}
where $\mathcal F$ and ${\mathcal F}^{{\text{ - }}1}$ denote the DFT and its inverse. ${{{\bf{k}}^{{\bf{ss}}}}}$ is the kernel correlation of the target appearance model $\bf{s}$ with itself. Each cyclically shifted training sample ${\bf{s}}_{w,h}$ actually consists of certain feature maps extracted from its corresponding image region. 

In the tracking process, a new image region $\bf{r}$ centered at the position of the last frame is cropped in the new frame. The position of the target is found in the maximum response of the output response map $f\left( {\bf{r}} \right)$.
\begin{equation}\label{tracking}
f\left( {\bf{r}} \right) = {{\mathcal F}^{ - 1}}\left( {{\mathcal F}\left( {{{\bf{k}}^{{\bf{sr}}}}} \right) \odot {\mathcal F}\left( {\bf{\alpha }} \right)} \right)
\end{equation} 
where $\odot$ is the element-wise product and ${{{\bf{k}}^{{\bf{sr}}}}}$ represents the kernel correlation of $\bf{s}$ and $\bf{r}$. 

Note that in KCF, $\bf{\alpha }$ in Eq. \ref{alpha} and the target appearance model $\bf{s}$ is updated continuously. The model will be corrupted when the object is occluded severely or totally missing and adapt to the wrong background or obstacle regions as shown in the third row of Fig. \ref{response}. This will lead to incorrect tracking results and missing the target in the following frames.  

\subsubsection{Analysis of the Response Map}
Severe occlusion and missing target are two significant challenges in visual tracking. As mentioned above, the KCF tracker cannot avoid the model corrupting due to the lack of the feedback from the tracking results. 

\begin{figure}[!t]
	\centering
	\subfigure [Normal]{\includegraphics[height=1.2in,width=1.6in,angle=0]{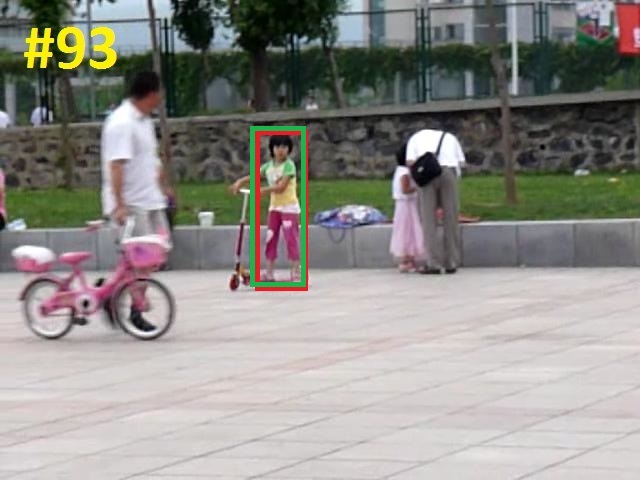}}
	\subfigure [PCE = 0.786]{\includegraphics[height=1.2in,width=1.6in,angle=0]{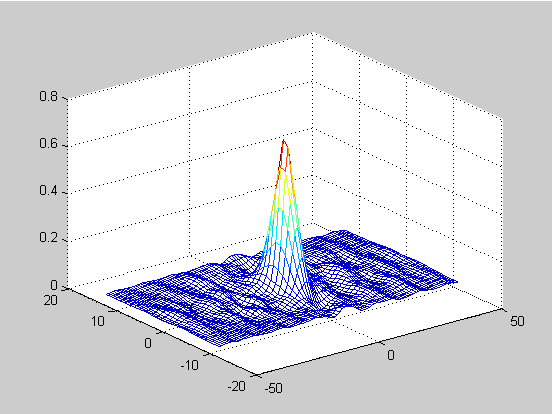}}
	\subfigure [Occluded]{\includegraphics[height=1.2in,width=1.6in,angle=0]{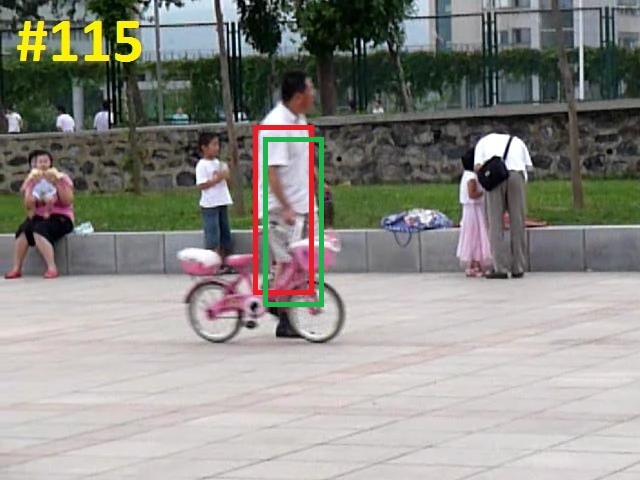}}
	\subfigure [PCE = 0.087]{\includegraphics[height=1.2in,width=1.6in,angle=0]{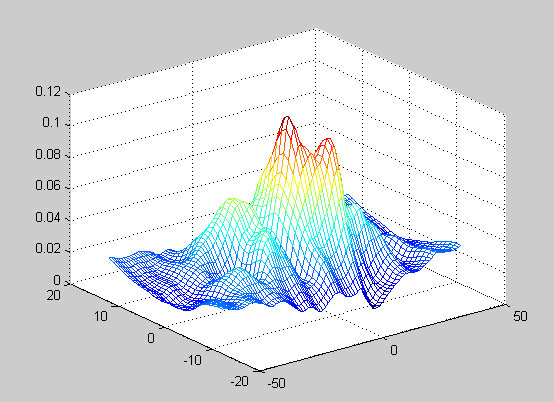}}
	\subfigure [Re-detected]{\includegraphics[height=1.2in,width=1.6in,angle=0]{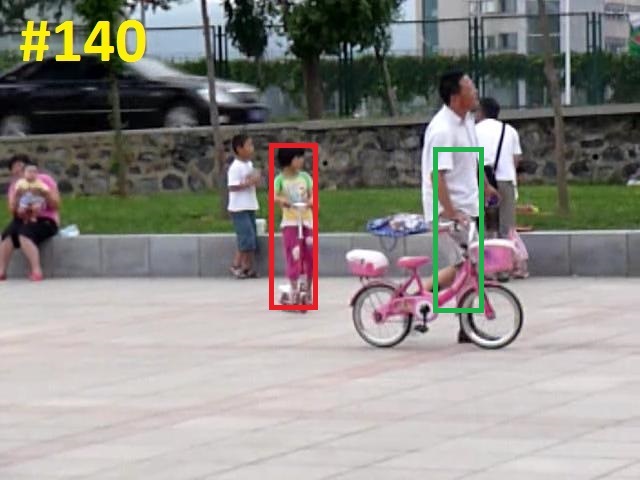}}
	\subfigure [PCE = 0.671]{\includegraphics[height=1.2in,width=1.6in,angle=0]{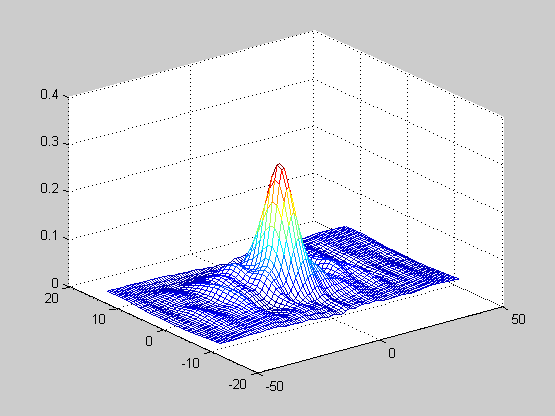}}	
	\caption{The first column shows the original frames from the vision sensors, the second column reveals the corresponding response maps. The red bounding box represents the found target of our method, while the green one denotes the tracking result of KCF tracker. When the girl is fully occluded, the corresponding response map will fluctuate intensely. By introducing the proposed criterion PCE, the target girl is re-detected again by our method and the response map returns to the normal pattern. However, the KCF tracker loses the target due to the model corrupting during the occlusion.}
	\label{response}
\end{figure}

The response map is the correlation response used to locate the position of target as in Eq. \ref{tracking}. It reveals the degree of confidence about the tracking results to some extent. The response map should have only one sharp peak and be smooth in all other areas when the detected target in the current frame is extremely matched to the correct target. The sharper the correlation peaks are, the better the location accuracy is. If the object is occluded severely or even missing, the whole response map will fluctuate intensely, resulting in a pattern that is significantly different from the normal response map as shown in Fig. \ref{response}. Instead of reporting a target regardless of the response map pattern, we propose a novel criterion for severe occlusion while remaining the advantages of KCF.

For correlation filter based classifier, the peak-to-sidelobe ratio (PSR) can be used to quantify the sharpness of the correlation peak. However, PSR is still not robust to partial occlusions \cite{liu2015real}. Therefore, we propose a novel criterion called peak-to-correlation energy (PCE) as described in Eq. \ref{PCE}. 
\begin{equation}\label{PCE}
PCE = \frac{{{{\left| {{y_{\max }}} \right|}^2}}}{{{E_y}}}
\end{equation} 
where ${\left| {{y_{\max }}} \right|}$ denotes the maximum peak magnitude, and the correlation response map energy $E_y$ is defined as Eq. \ref{Ey}. 
\begin{equation}\label{Ey}
{E_y} = \sum\limits_{w,h} {{{\left| {{y_{w,h}}} \right|}^2}} 
\end{equation} 

For sharper peak, i.e., the target apparently appearing in the visual field of the robot,  $E_y$ will get close to ${{{\left| {{y_{\max }}} \right|}^2}}$, thus PCE will approach to 1. Otherwise, PCE will approach to 0 if the object is occluded or missing. When the PCE is lower than a predefined threshold as shown in the second row of Fig. \ref{response}, the target appearance model and the filter model will not be updated.
\subsubsection{Scale Estimation}
When a robot tracks the target in front of it, the relative velocity of the robot and the target is changing all the time. And the size of the target in the image is varying according to the distance between the target and robot.

To handle scale variation ${s_{2D}}$ in the 2-d visual tracking process, we employ DSST \cite{danelljan2014accurate} algorithm. Firstly, the position of the object is determined by the learned translation filter with abundant features. Secondly, a group of windows with different scales are sampled around this position and correlated with the learned scale filter via coarse features. For each scale, the maximum value of its response map is measured as its matching score. The scale with the highest score is regarded as ${s_{2D}}$. At the meantime, the standard variance $\sigma _{2D}$ from ${s_{2D}}$ is calculated as the uncertainty of ${s_{2D}}$.  

We also consider the scale states calculated from the 3-D position estimations. At the $k$-th frame, the 3-D position ${{\bf{x}}_k}$ is estimated. Then we can get the 3-D positions of the head ${{\bf{x}}_{hk}}$ and feet ${{\bf{x}}_{fk}}$ by Eq.\ref{initial_def}(c)(d) as the height $h$ of the target is fixed during tracking. 

We can obtain ${{\bf{v}}_{hk}}$ and ${{\bf{v}}_{fk}}$ by projecting ${{\bf{x}}_{hk}}$ and ${{\bf{x}}_{fk}}$ into the image space through Eq. \ref{camera_measure:B}, where ${{\bf{v}}_{hk}}$ and ${{\bf{v}}_{fk}}$ are the $v$ axis values in the image space of head and feet, respectively. We assume that the scale variation of the height and width is synchronous. Then the scale variation from the 3-D position is obtained from
\begin{equation}\label{scale_3d}
{s_{3D}} = \left( {{v_{fk}} - {v_{hk}}} \right)/{v_{init}}
\end{equation} 
where $v_{init}$ is the initial height of the target calculated by Eq. \ref{initial:D}. Finally, the scale $s_k$ of the $k$-th frame is calculated using Eq.\ref{scale}.
\begin{equation}\label{scale}
{s_k} = \left( {\frac{{{s_{2D}}}}{{{\sigma _{2D}}}} + \frac{{{s_{3D}}}}{{{\sigma _{3D}}}}} \right)\frac{{{\sigma _{2D}}{\sigma _{3D}}}}{{{\sigma _{2D}} + {\sigma _{3D}}}}
\end{equation} 
where ${\sigma _{3D}} = \sqrt {{{\bf{P}}_k}\left( {3,3} \right)} $ is the uncertainty of $s_{3D}$ and ${{\bf{P}}_k}$ is the covariance matrix of the $k$-th estimated state.

\subsection{Ultrasonic Sensor Tracking Model}
Traditional sonar array sensors use time-of-flight (TOF) and triangulation to find the relative location of a target with respect to the source. In the proposed tracking system, the Gaussian Process Regression (GPR) techniques are used in sonar sensor tracking model to obtain the range information and improve the predicted accuracy of the tracking target \cite{su2014ultrasonic}.  

The active sonar emitter array which consists of three sonar sensors is designed as a human carrying Portable User Device (POD). The corresponding passive sensor array with four sonar units is attached equally spaced in front of the robot. When the RF module on the robot receives the RF signal from the POD, it will start a timer. Then the time lapsed from when the timer starts until all the sonar units measure an incoming signal is the corresponding TOF.

For each sonar unit, GPR model trained with real data is built to predict sensor reading with corresponding covariance at a certain $\left( {{x_{Uk}},{y_{Uk}}} \right)$ location, where the subscript $Uk$ denotes the $k$-th ultrasonic state. The final posterior probability for prediction is calculated by the Eq.\ref{GPR}. The position with highest probability is chosen as the predicted location.
\begin{equation}\label{GPR}
P\left( {\left. {{x_{Uk}},{y_{Uk}}} \right|{U_1},{U_2},{U_3},{U_4}} \right) = 
\frac{{\prod\limits_{i = 1}^4 {P\left( {{U_i}\left| {{x_{Uk}},{y_{Uk}}} \right.} \right)} P\left( {{x_{Uk}},{y_{Uk}}} \right)}}{{P\left( {{U_1},{U_2},{U_3},{U_4}} \right)}}
\end{equation}
where ${P\left( {{U_i}\left| {{x_{Uk}},{y_{Uk}}} \right.} \right)}$, $i=1,...,4$ are the learned GPR models for 4 sonar units. ${P\left( {{x_{Uk}},{y_{Uk}}} \right)}$ is the prior inference for unknown position ${\left( {{x_{Uk}},{y_{Uk}}} \right)}$ and is assumed as a uniform distribution. As the $U_1$ to $U_4$ are the actual observations, ${P\left( {{U_1},{U_2},{U_3},{U_4}} \right)}=1$.

Transform to the $k$-th system state ${{\bf{x}}_k}{\text{ = }}{\left[ {{x_k},{y_k},{z_k}} \right]^T}$ in the robot coordinate, the expected measurement ${{\bf{h}}_U}\left( {{{\bf{x}}_k}} \right)$ of the ultrasonic sensor is denoted as
\begin{subequations}\label{sonar_measure}
	\begin{align}
	{{\bf{h}}_U}\left( {{{\bf{x}}_k}} \right) &= {\left[ {{x_{Uk}},{y_{Uk}}} \right]^T}\label{sonar_measure:A} \\
	{x_{Uk}} &= \sqrt {{x_k}^2 + {z_k}^2} \label{sonar_measure:B} \\
	{y_{Uk}} &= {y_k}\label{sonar_measure:C}	
	\end{align}
\end{subequations}

\subsection{Multi-Sensor Fusion}
A standard EKF approach is utilized to fuse the measurements obtained from the ultrasonic sensor and the vision sensor. As the update frequencies of the two sensors are different, the fusion algorithm will be triggered whenever any of them is updated. We adopt such a method to sequentially process the multiple, heterogeneous measurements arriving in an asynchronous order \cite{durrant2008multisensor}.
\subsubsection{Prediction Step}
As we have no knowledge of the target motion, a random walk or a constant velocity model can be used to predict the target location in the robot coordinate. In the case of random walk model,
\begin{subequations}\label{prediction}
	\begin{align}
	{{\bf{x}}_k} &= {{\bf{x}}_{k - 1}} \label{prediction:A} \\
	{{\bf{P}}_k} &= {\bf{G}}{{\bf{P}}_{k - 1}}{{\bf{G}}^T} + {{\bf{R}}_k} \label{prediction:B} \\
	{{\bf{R}}_k} &= {\bf{R}}\left( {{t_k} - {t_{k - 1}}} \right)\label{prediction:C}	
	\end{align}
\end{subequations}

where ${\bf{P}}_k$ is the covariance matrix, $\bf{G}$ is the Jacobian of Eq. \ref{prediction:A} (3 by 3 identity matrix in the random walk model). $\bf{R}_k$ is the motion noise during the time step $\left( {{t_k} - {t_{k - 1}}} \right)$, so it is proportional to the $\left( {{t_k} - {t_{k - 1}}} \right)$ by a constant noise level $\bf{R}$.
\subsubsection{Correction Step}
Whenever any measurement is available, the system state is updated by
\begin{subequations}\label{correction}
	\begin{align}
	{{\bf{K}}_{*k}} &= {{\bf{P}}_k}{\bf{H}}_{*k}^T{\left( {{{\bf{H}}_{*k}}{{\bf{P}}_k}{\bf{H}}_{*k}^T + {{\bf{Q}}_{*k}}} \right)^{ - 1}} \label{correction_sonar:A} \\
	{{\bf{x}}_k} &= \;{{\bf{x}}_k} + {{\bf{K}}_{*k}}\left( {{{\bf{z}}_{*k}} - {{\bf{h}}_*}\left( {{{\bf{x}}_k}} \right)} \right)\label{correction:B} \\
	{{\bf{P}}_k} &= \left( {{\bf{I}} - {{\bf{K}}_{*k}}{{\bf{H}}_{*k}}} \right){{\bf{P}}_k}\label{correction:C}	
	\end{align}
\end{subequations}

where the $*$ in the subscript stands for either ultrasonic $\left( U \right)$ or camera $\left( C \right)$ measurement, we collectively call it measurement next. ${{{\bf{h}}_*}\left( {{{\bf{x}}_k}} \right)}$ is the sensor model which provides the predicted measurement. The camera sensor model ${{{\bf{h}}_C}\left( {{{\bf{x}}_k}} \right)}$ is define in Eq.\ref{camera_measure} and the sonar sensor model ${{{\bf{h}}_U}\left( {{{\bf{x}}_k}} \right)}$ is defined in Eq.\ref{sonar_measure}. ${{{\bf{H}}_{*k}}}$ is the Jacobian matrix of ${{{\bf{h}}_*}\left( {{{\bf{x}}_k}} \right)}$. ${{{\bf{z}}_{*k}}}$ is the actual sensor measurement. ${{{\bf{z}}_{Ck}}}$ is estimated from the visual tracking algorithm by Eq.\ref{tracking} and ${{{\bf{z}}_{Uk}}}$ is the mean predicted by GPR in Eq.\ref{GPR}. ${{{\bf{Q}}_{*k}}}$ is the measurement noise. For sonar sensor, ${{{\bf{Q}}_{Uk}}}$ is the covariance matrix from Gaussian process. For camera sensor, ${{{\bf{Q}}_{Ck}}}$ is defined as 
\begin{equation}\label{Q-Ck}
{{\bf{Q}}_{Ck}} = \left[ {\begin{matrix}
	{\frac{{0.002}}{{PCE}}} & 0  \\ 
	0 & {\frac{{0.002}}{{PCE}}}  \\ 
	
	\end{matrix} } \right]
\end{equation}

 So the input from the camera sensor will be ${{{\bf{z}}_{Ck}}}$ and ${{{\bf{Q}}_{Ck}}}$, both coming from the visual tracking algorithm. The input from the sonar sensor will be ${{{\bf{z}}_{Uk}}}$ and ${{{\bf{Q}}_{Uk}}}$, both coming from GPR.

\section{Experiments}
\begin{figure}[!t]
	\centering
	\subfigure [Simulation]{\includegraphics[height=1.1in,width=1.6in,angle=0]{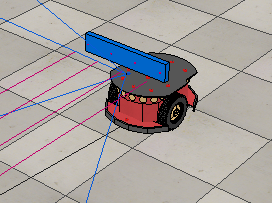}} \label{car:A}
	\subfigure [GUTS]{\includegraphics[height=1.1in,width=1.6in,angle=0]{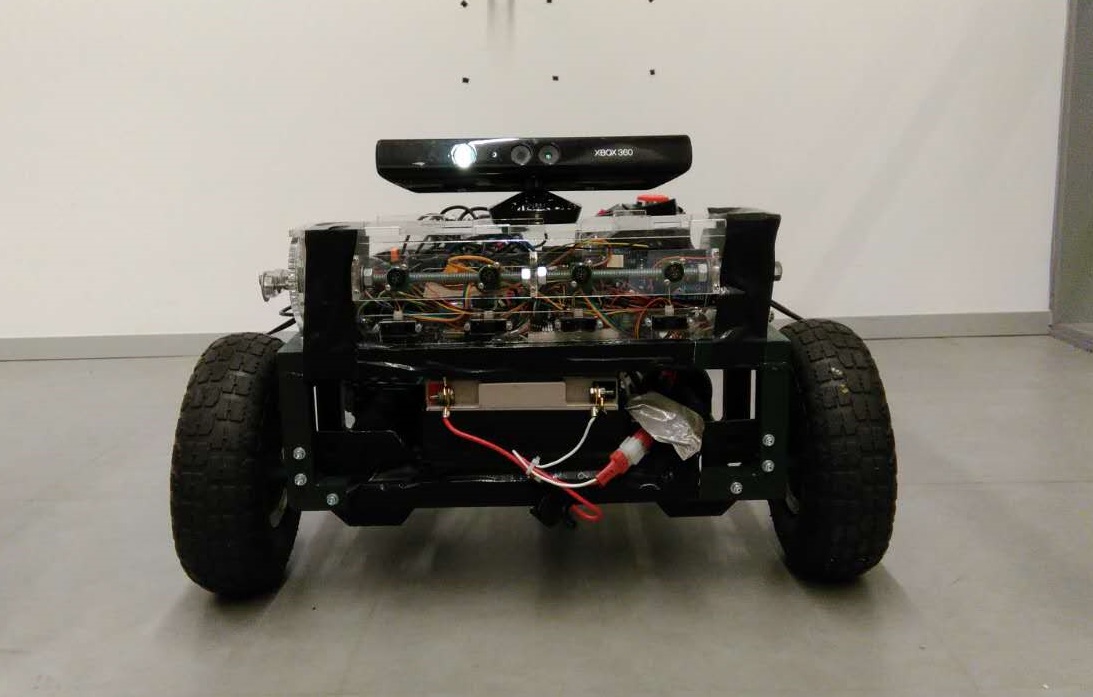}}\label{car:B}
	\caption{(a) Simulated and (b) Physical robot platforms. The Kinect on the GUTS is used to obtain the ground truth of the human as well as the monocular camera sensor.}
	\label{car}
\end{figure}

To evaluate the performance of our multi-sensor 3-D human tracking system, various experiments were carried out both simulation environments and real world scenarios. The simulation is done by a robot simulator named Virtual Robot Experimentation Platform (V-REP) which is used for fast prototyping and verification to validate the accuracy of the proposed tracking system. The actual robot platform is named Garden Utility Transportation System (GUTS), shown in Fig.\ref{car}b. It is a differential mobile robot system fitted with an auto-tipping mechanism \cite{su2014ultrasonic}. The detailed tracking processes of all the experiments are shown in our video demo.
\begin{table*}[!t]
	\centering
	\caption{The arithmetic mean error in three axis. $S$ is short for simulation, $G$ for GUTS, $I$ for indoor, $O$ for outdoor, $C$ for camera sensor, and $U$ for ultrasonic sensor.}
	\begin{tabular}{||c|c|c|c|c|c|c||} 
		\hline
		Axis & SI(C+U)  &	SO(C+U) & 	SO(C)& 	SO(U) &
		GI(C+U) & GO(C+U) \\ 
		\hline\hline
		x(m)	& 0.144	 & 0.07521	& 1.1650	& 0.1413	& 0.168	& 	0.2045\\ 
		\hline
		y(m)	&0.1062		&	0.03776	&	0.1169	&	0.1364	&	0.1091		&	0.1262\\
		\hline
		z(m)&	0.1085&	0.1167&	0.2679&	0.4366&	0.116&	0.1231\\
		
		\hline
	\end{tabular}
	\label{table:1}
\end{table*}
\subsection{Experimental Setup}
All our experiments are performed using MATLAB R2015a on a 3.2 GHz Intel Core i7 PC with 16 GB RAM. The Robot Operating System (ROS) has been employed as the software framework for the GUTS platform, linked to MATLAB via the MATLAB-ROS bridge package. The visual tracking algorithm runs at an average speed of 25 FPS and the GPR prediction of sonar runs at about 5 Hz. The ultrasonic sensor tracking algorithm remains the same as described in our previous work \cite{su2014ultrasonic}. The setup of the visual tracking algorithm is introduced below. 
\begin{figure}[!t]
	\centering
	\subfigure [Indoor]{\includegraphics[height=1.2in,width=1.6in,angle=0]{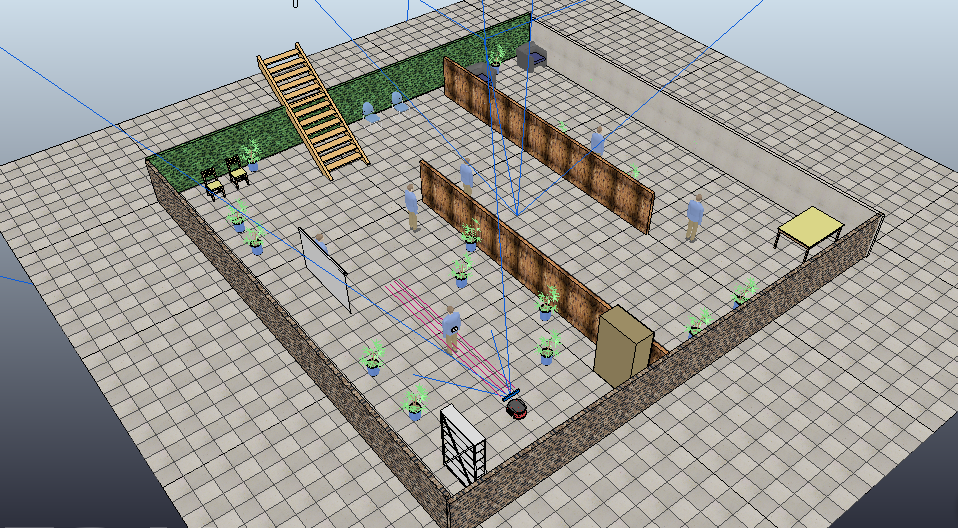}} 
	\subfigure [Outdoor]{\includegraphics[height=1.2in,width=1.6in,angle=0]{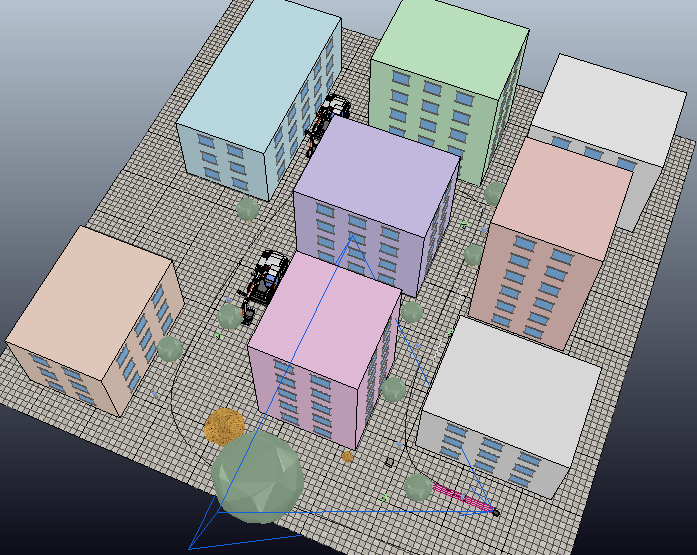}}
	\caption{Simulation scenes.}
	\label{Sim_scenes}
\end{figure}

Multichannel features based on Histogram of Oriented Gradient (HOG) \cite{dalal2005histograms} with a cell size of 4 pixels, as well as color names (CN) \cite{danelljan2014adaptive} with the first 2 dimensions, are used in our method. The threshold of the PCE criterion is set to 0.2 from experiments. When PCE is larger than 0.2, the target appearance and the filter models are updated. Otherwise, the target is perceived as occluding or missing, so the target appearance and the filter models are not updated. The regularization parameter $\lambda$ in Eq.\ref{squared_error} is set to 0.0001.

In order to demonstrate the performance of the proposed 3-D tracking system, we test it in the simulation experiments and the GUTS robotic platform in both indoor and outdoor environments. To illustrate that under normal walking paces and patterns the proposed tracking system is able to effectively track the target people, we apply a simple proportional controller in translation and orientation velocity to make the robot track automatically.
\subsection{Experimental Results}
The simulated robotic platform is depicted in Fig.\ref{car}a. A passive sensor array with four sonar receiver units is mounted equally spaced in front of the robot. The camera is fixed below the sonar array. Indoor scene is constructed as an office room with many people walking inside it as shown in Fig.\ref{Sim_scenes}a. Outdoor scene in Fig.\ref{Sim_scenes}b is built with plenty of buildings, trees, vehicles and people to imitate a city area. 

To validate the necessity of the two sensors, we compare the tracking results with individual sensor respectively only in the simulation outdoor scene for safety. Without the monocular camera sensor, the estimated accuracies of $z$ axis is dramatically reduced. Without the sonar sensor, the estimation of $x$ axis is incorrect due to the lack of information in this dimension. The tracking result is imponderable when the visual tracking algorithm loses the target. The corresponding mean errors are shown in TABLE. \ref{table:1}.

\begin{figure}[!t]
	\centering
	\subfigure [Indoor]{\includegraphics[height=1.4in,width=1.6in,angle=0]{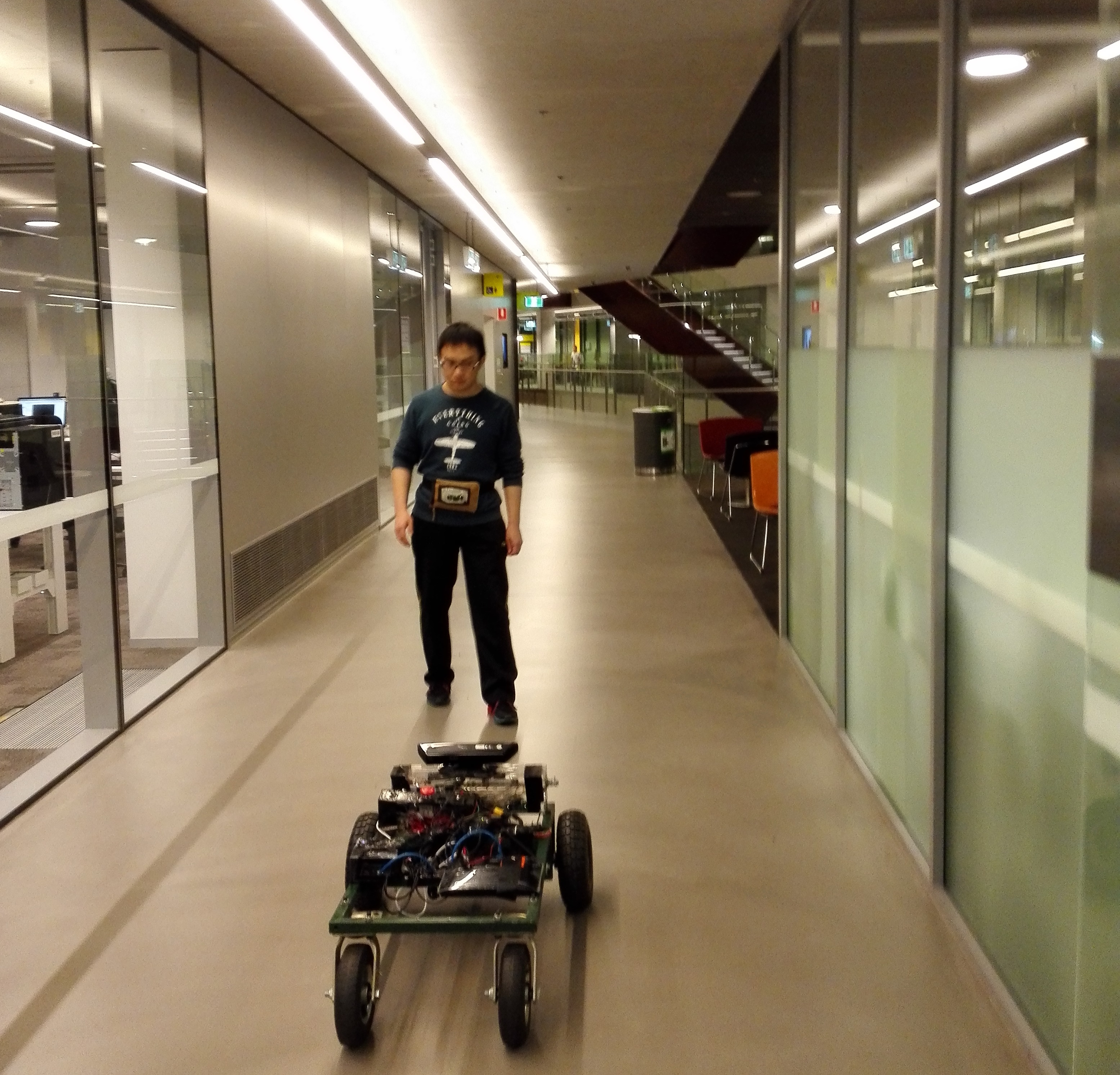}} 
	\subfigure [Outdoor]{\includegraphics[height=1.4in,width=1.6in,angle=0]{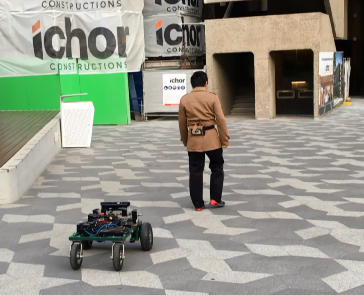}}
	\caption{Real world scenarios.}
	\label{scenes}
\end{figure}

To evaluate the performance of the proposed 3-D tracking system in reality, more experiments were conducted with the GUTS platform as shown in Fig. \ref{car}b. We introduce the skeleton tracking of Kinect XBOX 360 to collect the ground truth of the 3-D positions of the target during tracking process through the OpenNI tracker in ROS. The position of the waist in the skeleton is regarded as the true position of the sonar POD carried by the person. Simultaneously, the RGB camera on the Kinect is used as our monocular camera sensor. As shown in Fig.\ref{scenes}, the indoor experiment is performed in the common corridor of our laboratory while the outdoor experiment is conducted outside the main building of University of Technology Sydney.

\begin{figure*}[!t]
	\centering
	\subfigure [Simulated Indoor Scene]{\includegraphics[height=1in,width=1.7in,angle=0]{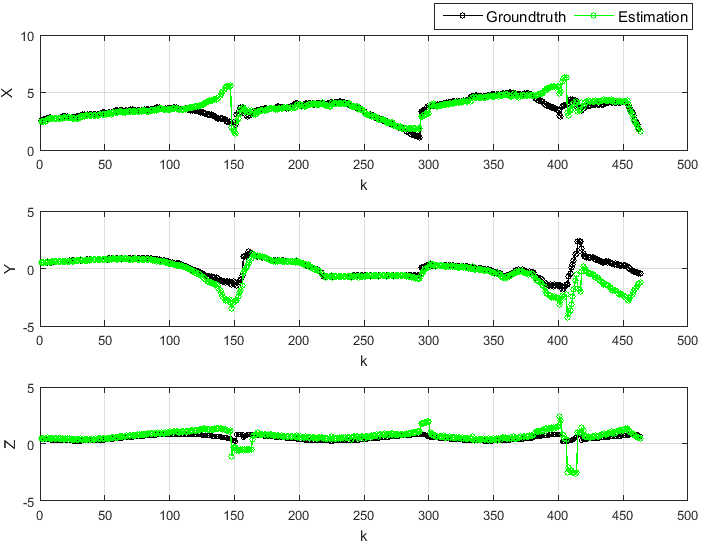}}
	\subfigure [Simulated Outdoor Scene]{\includegraphics[height=1in,width=1.7in,angle=0]{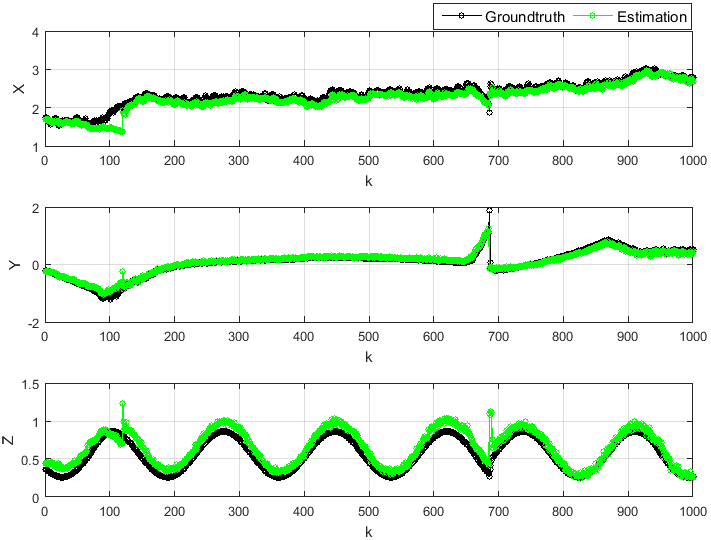}}
	\subfigure [Actual Indoor Scene]{\includegraphics[height=1in,width=1.7in,angle=0]{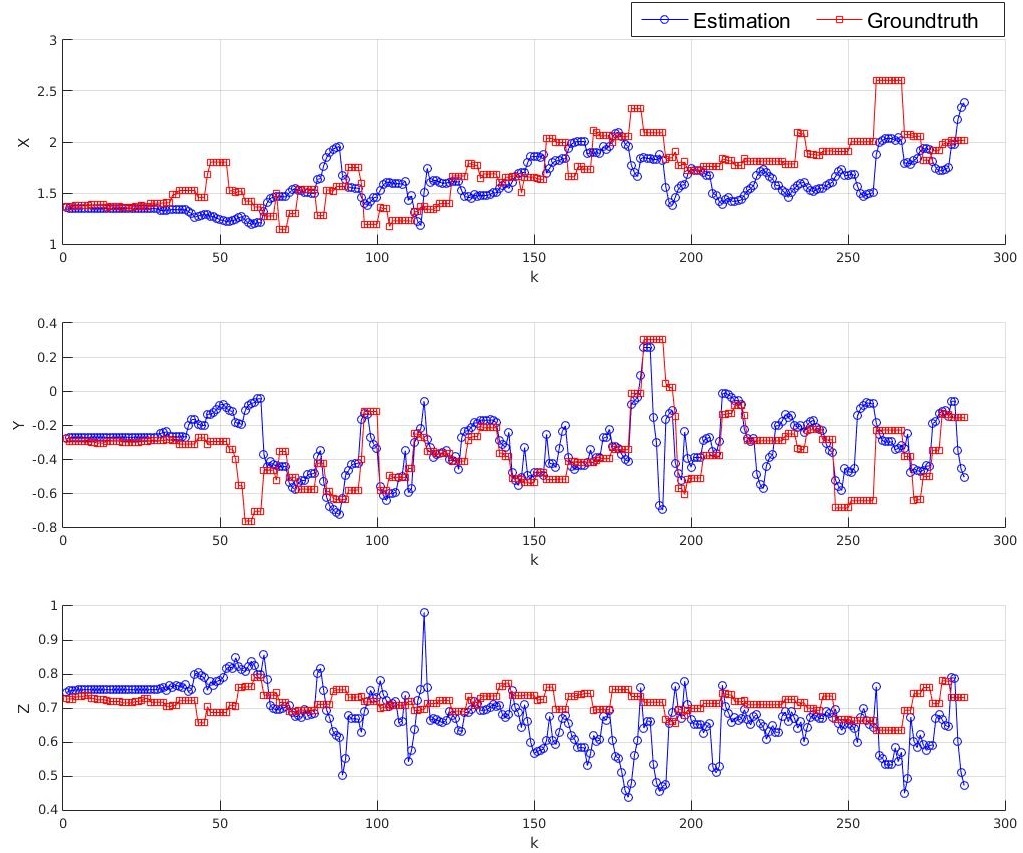}}
	\subfigure [Actual Outdoor Scene]{\includegraphics[height=1in,width=1.7in,angle=0]{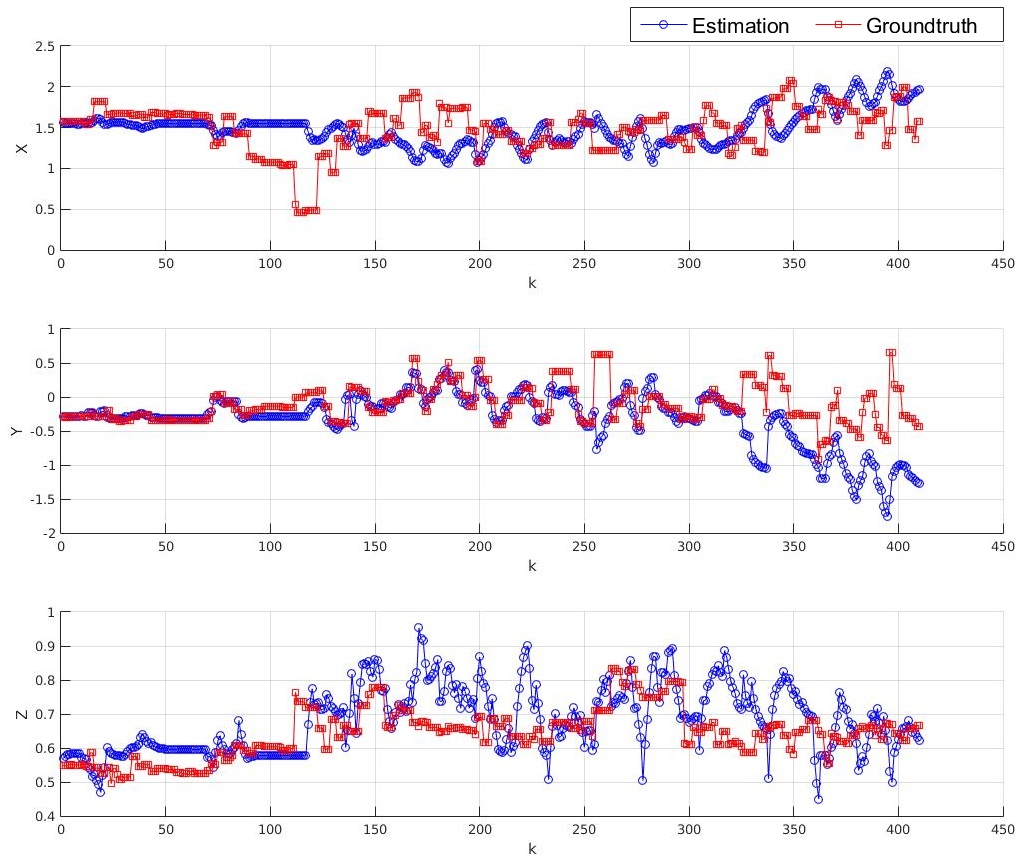}}
	\caption{3-D tracking results of simulation and actual experiments. Results are best viewed on high-resolution displays.}
	\label{Results}
\end{figure*}

There are many challenges in these scenes such as illuminate variations, scale variations, part occlusions, severe occlusions, background clutters and object missing. The target people is walking with the variations in all three axes to make the 3-D estimation more challenging. We show these challenges in our video demo. The quantitative 3-D tracking results are shown in Fig. \ref{Results}a,b for simulation scenes, Fig. \ref{Results}c,d for real-world scenarios.

The results illustrate a great performance of our method. As shown in Fig. \ref{Results}a,b for the indoor and outdoor scenes of the simulation experiments, the black lines represent the ground truth of the target motions in three axes, the green lines denote the estimation of the proposed tracking system. It can be observed that the tracking errors is markedly small since the two lines are closed to each other in all three axes. Also, in Fig. \ref{Results}c,d for the indoor and outdoor experiments in GUTS, the red lines show the ground truth of the target motions while the blue ones denote the estimation of our method. We calculate the mean errors of all the experiments in TABLE \ref{table:1}.


\section{Conclusion}
In this paper, we address the problem of accurately estimating the 3-D position of the target around the mobile robot for tracking purposes, in both indoor and outdoor environments. Our approach fuses the partial location estimations from a monocular camera and an ultrasonic array. To improve the robustness of the tracking system, a novel criterion in the visual tracking model is introduced to overcome the problems of occlusions, scale variation, targets missing and re-detection. The ultrasonic sensor is used to provide the range based location estimation. Information from two heterogeneous sources is processed with EKF sequentially to handle their different update rates. The estimated 3-D information is further exploited to improve the scale accuracy. The proposed approach is implemented and tested in both simulation and real-world scenarios. As the evaluation results show, the proposed algorithm is able to produce stable, accurate and robust 3-D position estimations of the target in real-time.
\section*{Acknowledge}
This work was supported in part by the National Natural Science Foundation of China under Grant U1509210 and Grant U1609210, and in part by the Natural Science Foundation of Zhejiang Province under Grant LR13F030003.





%
\bibliographystyle{ieeetr}
\bibliography{reference}

\end{document}